\documentclass[letterpaper, 10 pt, journal, twoside]{IEEEtran}

\usepackage{graphicx}
\usepackage{cite}
\usepackage{picinpar}
\usepackage{amsmath}
\usepackage{url}
\usepackage[latin1]{inputenc}
\usepackage{colortbl}
\usepackage{soul}
\usepackage{multirow}
\usepackage{pifont}
\usepackage{color}
\usepackage{alltt}
\usepackage{hyperref}
\usepackage{amssymb}
\usepackage{enumerate}
\usepackage{siunitx}
\usepackage{breakurl}
\usepackage{epstopdf}
\usepackage{pbox}
\usepackage[ruled,linesnumbered]{algorithm2e}
\usepackage{booktabs} 
\usepackage[caption=false,font=footnotesize]{subfig}
\usepackage{bold_letter}

\usepackage{bm}

\ifCLASSINFOpdf
\else
\fi

\begin{document}
\title{
LaSeSOM: A Latent and Semantic Representation\\ Framework for Soft Object Manipulation}

\author{
Peng Zhou, \IEEEmembership{Student Member,~IEEE}, Jihong Zhu, \IEEEmembership{Member,~IEEE}, Shengzeng Huo, \IEEEmembership{Student Member,~IEEE}, \\and David Navarro-Alarcon, \IEEEmembership{Senior Member,~IEEE}
\thanks{Manuscript received December 24, 2020; Revised March 21, 2021; Accepted April 13, 2021. This paper was recommended for publication by Editor Hong Liu upon evaluation of the Associate Editor and Reviewers' comments.
This work is supported by the Research Grants Council under Grant 14203917, in part by the PROCORE-France/Hong Kong Joint Research Scheme under Grant F-PolyU503/18, in part by the Key-Area Research and Development Program of Guangdong Province 2020 under project 76, in part by the Jiangsu Industrial Technology Research Institute Collaborative Research Program Scheme under Grant ZG9V, and in part by PolyU under Grants 252047/18E, ZZHJ, and UAKU. \emph{(Corresponding author: David Navarro-Alarcon.)}}
\thanks{P. Zhou, S. Huo and D. Navarro-Alarcon are with The Hong Kong Polytechnic University, KLN, Hong Kong (e-mail: jeffery.zhou@connect.polyu.hk; kyle-sz.huo@connect.polyu.hk; dna@ieee.org)
}
\thanks{J. Zhu is with Delft University of Technology, Mekelweg 2, 2628CD, The Netherlands. (e-mail: j.zhu-3@tudelft.nl)
}%
\thanks{Digital Object Identifier (DOI): see top of this page.}
}

\markboth{IEEE Robotics and Automation Letters. Preprint Version. April, 2021}
{Zhou \MakeLowercase{\textit{et al.}}: LaSeSOM: A Latent and Semantic Representation Framework for Soft Object Manipulation} 

\maketitle

\begin{abstract}
Soft object manipulation has recently gained popularity within the robotics community due to its potential applications in many economically important areas. 
Although great progress has been recently achieved in these types of tasks, most state-of-the-art methods are case-specific; They can only be used to perform a single deformation task (e.g. bending), as their shape representation algorithms typically rely on ``hard-coded'' features.
In this paper, we present LaSeSOM, a new feedback latent representation framework for semantic soft object manipulation. 
Our new method introduces internal latent representation layers between low-level geometric feature extraction and high-level semantic shape analysis; This allows the identification of each compressed semantic function and the formation of a valid shape classifier from different feature extraction levels. 
The proposed latent framework makes soft object representation more generic (independent from the object's geometry and its mechanical properties) and scalable (it can work with 1D/2D/3D tasks). 
Its high-level semantic layer enables to perform (quasi) shape planning tasks with soft objects, a valuable and underexplored capability in many soft manipulation tasks.
To validate this new methodology, we report a detailed experimental study with robotic manipulators.
\end{abstract}

\begin{IEEEkeywords}
Bimanual Manipulation; Representation Learning; Shape Deformation Planning; Latent Space and Manifolds; Geodesic Interpolation.
\end{IEEEkeywords}

\IEEEpeerreviewmaketitle

\section{Introduction}
\IEEEPARstart{R}{ecent} studies have shown that the manipulation of soft objects is crucial and indispensable to achieve high autonomy in robots \cite{amor2014special}. 
Although great progress has been recently achieved, the \emph{feedback} manipulation of soft objects is still a challenging research question. 
The implementation of these types of advanced manipulation capabilities is complicated by various issues. 
Amongst the most important is the difficulty in characterizing the feedback shape of a soft object.
Our aim in this work is to develop new data-driven methods that can quantitatively describe deformable shapes. 

\begin{figure}[htbp]
\centering
\includegraphics[width=\linewidth]{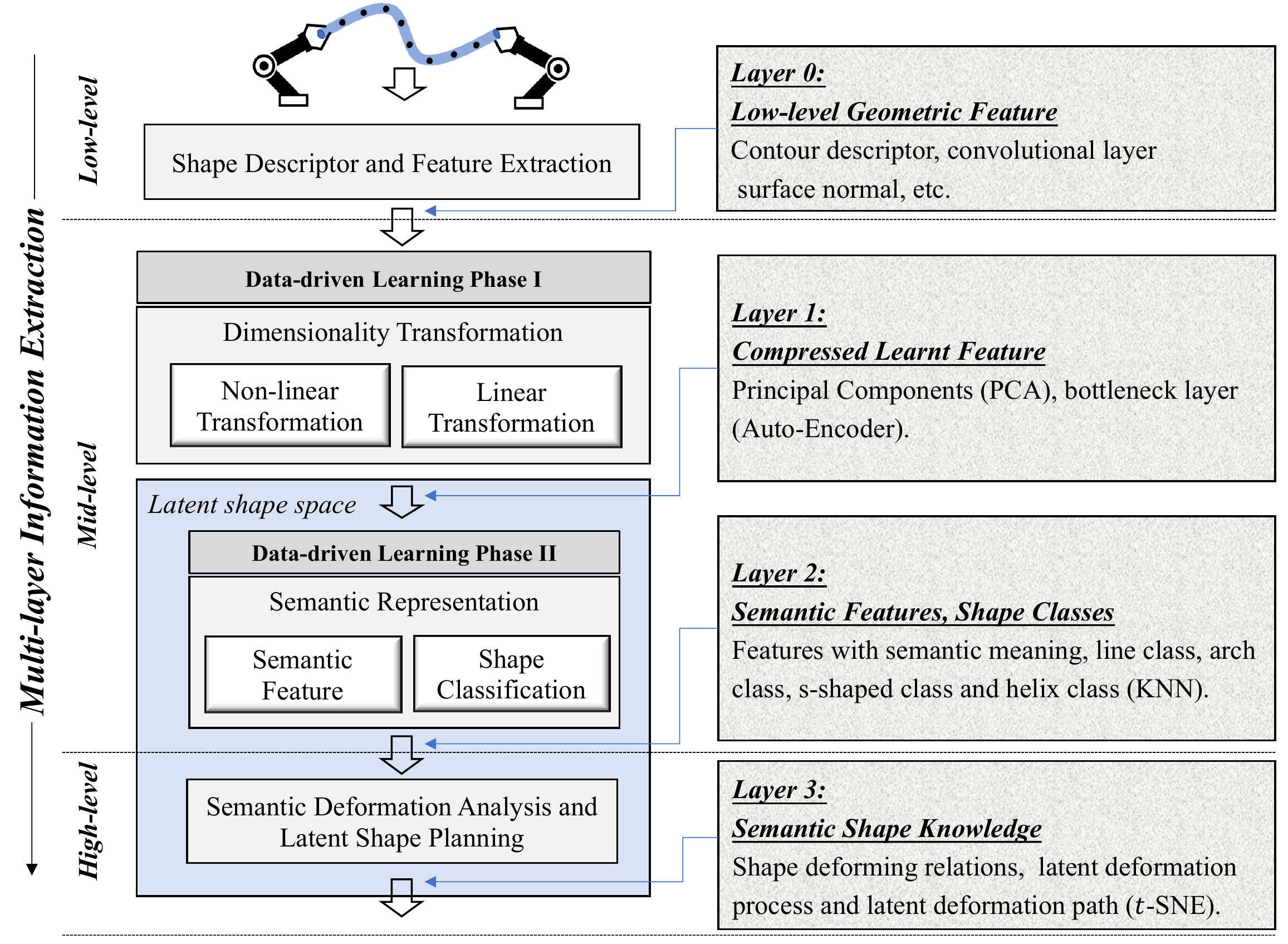}
\caption{Conceptual representation of the proposed framework --- LaSeSOM that fully describes and represents the soft objects for bimanual manipulation tasks from four layers, namely, the low-level geometric feature layer, compressed learnt feature layer, semantic features and shape classes layer, and semantic shape knowledge layer.}
\label{fig_framework_entire}
\end{figure}

Hirai \cite{Journals:Hirai2000} first demonstrated how feedback controls could deform a soft object into a desired 2D shape. This early work is a clear example of a \emph{shape representation} based on points \cite{zerui2018_iros} (simple, but cannot generalize). Other classical methods are based on geometric features e.g. angles, curvatures, catenaries \cite{navarro2014visual,LARANJEIRA2020107018}; Its disadvantage is that they are case-specific, thus, can only be used to perform a single shaping action. 
Some works have addressed this issue by developing generic representations that only require sensory data. For example, \cite{dna2018_tro,Proceedings:Zhu2018}, and \cite{Journals:Hu2018} characterize shapes using Fourier series and feature histograms; These methods, however, create very large feature vectors, which may not be the most efficient feedback metric. A more effective solution is to automatically compute generic feedback features (e.g. as in direct visual servoing \cite{Journals:Collewet2011,Journals:Marchand2019}) and combine them with dimension reduction techniques, as in e.g. \cite{quantifyShapes,zhu2020vision}.
Data-driven based shape analyses \cite{xu2016data}, \cite{zhang2008deformation} have gained in popularity as it offers a useful alternative to model-based approaches. An increasing amount of research have focus on different-level segmentation and shape classifications (see \cite{golovinskiy2009consistent}, \cite{sidi2011unsupervised}, and \cite{van2013co}). 
However, these methods purely depend on the designed end-to-end pipeline which ignores the semantic meaning of internal features and thus
 failing to interpret the entire analytical process. 
 Therefore, latest applications started to examine attribute-based approaches, such as binary attributes \cite{tao2009skyfinder}, relative attributes \cite{parikh2011relative}, and semantic image color palette editing \cite{laffont2014transient}. 
  Several works \cite{leifman2005semantic}, \cite{attene2009characterization} further combine shape analysis and semantic attributes for a in-depth deformation analysis.


Latent space approaches have recently achieved many successful results in image analysis \cite{hoff2002latent}, due to its capability to encode high-dimensional data into a meaningful internal representation.
By using concise low-dimensional latent variables and highly flexible generators, a latent space allows us to generate new data samples on data space. 
In this manner, a deformation planning problem of soft objects can be solved in a novel way by constructing a feasible sequence of deformable shapes in latent space. 
However, many works \cite{achlioptas2018learning} have adopted a linear interpolation in remapping the latent variables back to data space, which could cause serious distortions on the generated samples for a shape planning scenario.
For example, consider a generator $g$ and a latent variable $\mathbf{z}$ with two infinitesimal shifts $\delta_1$ and $\delta_2$, then the distance with Taylor's expansion \cite{arvanitidis2018latent} is formulated by:
\begin{equation}\small
	\left\|g\left(\mathbf{z_0}+\delta_{1}\right)-g\left(\mathbf{z_0}+\delta_{2}\right)\right\|^{2} \\=\left(\Delta_{12}\right)^{\top}\left(\mathbf{J}_\mathbf{z_0}^{\top} \mathbf{J}_{\mathbf{z_0}}\right)\left(\Delta_{12}\right) 
\end{equation}
for $\mathbf{J}_{\mathbf{z_0}}=\left.\frac{\partial g}{\partial \mathbf{z}}\right|_{\mathbf{z}=\mathbf{z_0}}$ and $\Delta_{12} = \delta_1 - \delta_2$, which indicates that the normal distance in $\mathcal{Z}$ space changes locally as it is determined by the local Jacobian.
Consequently, seeking the shortest curve along a curved surface, a manifold, manifold is a more reasonable way to compute the interpolation and generate undistorted samples. 

As a feasible solution to these problems, we present a general data-driven representation framework --- \textbf{LaSeSOM} for semantic soft object manipulation depicted in Fig. \ref{fig_framework_entire}, which is composed of three layers: A low-level soft object geometric shape processing, a mid-level data-driven representation learning, and a high-level semantic shape analysis. 
The paper's main contributions are summarized as follows:
\begin{itemize}
  \item An effective representation framework for soft object analysis during manipulation tasks.
  \item A novel semantic analysis approach for soft object manipulation tasks.
  \item A solution for shape planning with a geodesic path-based interpolation algorithm in the latent space.
\end{itemize}

The rest of this paper is organized as follows. Section \ref{sec:methods} presents the representation models. Section \ref{sec:results} shows the experimental results. Section \ref{sec:con} gives final conclusions.

\section{Methods}\label{sec:methods}
In \textbf{LaSeSOM}, we first introduce two shape features extracted from two data formats for shape description (marker points and point clouds), and then two dimensionality transformation techniques for building latent space.
With this latent space, we design several semantic analysis algorithms to describe soft object deformations and solve the deformation planning problem.

\subsection{Shape Feature}
In order to apply this framework into various soft object manipulation tasks, two typical data formats are selected to depict the soft object shape. One is the ordered marker point data in the format of a set of ordered 3D points that is widely used in the motion tracking system, and the other is a popular point cloud data to represent a geometric shape surface via a set of large quantities of unordered 3D points in a Euclidean space. 
Formally, Let  $\mathcal{S}=\left\{\mathcal{S}_{1}, \ldots, \mathcal{S}_{p} \mid \mathcal{S}_{i} \in \mathbb{R}^{q\times3}\right\}$ be set of a complete soft object deformation, and $\mathcal{S}_{i}$ denotes the $i$-th shape during the deformation process. Using $q$ marker points,
$\mathcal{S}_i=\left\{\bf{x}_{1}, \ldots, \bf{x}_{q} \mid \bf{x}_i \in  \mathbb{R}^3 \right\}$ can be determined by an ordered 3D points set.
Consequently, the entire deformation can represented as a shape matrix $\mathbf{X}_{in} \in \mathbb{R}^{p \times 3q}$,
where the coordinates of the markers have been fatten so each row with $q$ markers has $3q$ features and the number of total shapes during this deformation is denoted by $p$. 
To approximate the contour composed of 3D marker points, Fourier approximation \cite{zhang2002comparative} is selected considering that this descriptor can depict the shape with arbitrary precision. However, 
instead of using its common 2D modeling form, we expand this descriptor into a 3D configuration as below:
\begin{equation}\small
\label{equ:3dfourier}
\begin{split}
x(l) = a_0 + \sum_{N}^{n=1}(a_ncos(wnl) + b_nsin(wnl)) \\
y(l) = c_0 + \sum_{N}^{n=1}(c_ncos(wnl) + d_nsin(wnl)) \\
z(l) = e_0 + \sum_{N}^{n=1}(e_ncos(wnl) + f_nsin(wnl))
\end{split}
\end{equation}
where $a_0$, $c_0$, and $e_0$ are the bias components of the Fourier descriptor with a frequency of 0, and $l$ is a same length that periodically circles along the entire length of soft object denoted by $L$. The coefficients of the $n$-th harmonic are denoted by $a_n, b_n, \ldots, f_n$, which can be solved with expressions in \cite{zhang2002comparative} to constitute the description of the shape. 

A deformable shape $\mathcal{S}_i$ can also be represented as a point cloud data $\mathcal{P}_i$. With farthest point sampling algorithm used in PointNet++ \cite{qi2017pointnet}, the raw point cloud can be sampled into $\mathcal{P}^{'}_{i}$ with a fixed input size $3N$, where $N$ is the resolution of the resampled point cloud, which means is the total number of points in this point cloud. Thus, given a point cloud $\mathcal{P}^{'}$, the input shape matrix can be represented as $\mathbf{X}_{in} \in \mathbb{R}^{N \times 3}$.
The feature extraction process follows the design principle of PointNet \cite{qi2017pointnet0}: increasing the features with convolutional 1D layers (thus, each point in $\mathcal{P}^{'}$ can be encoded independently); After the convolutions is connected a ``symmetric" and permutation-invariant function (e.g. a max pooling) to generate a joint feature representation in a size of $1 \times N$.
In this paper, we select the Chamfer(pseudo)-distance (CD) as the permutation-invariant metric for comparing unordered point sets. Given two point cloud set $\mathcal{P}_i$ and $\mathcal{P}_j$, this metric measures the squared distance between corresponding nearest neighbors in different sets:
\begin{equation}
	d_{CD}\left(\mathcal{P}_{i}, \mathcal{P}_{j}\right)=\sum_{x \in \mathcal{P}_{i}} \min _{y \in \mathcal{P}_{j}}\|x-y\|_{2}^{2}+\sum_{y \in \mathcal{P}_{j}} \min _{x \in \mathcal{P}_{i}}\|x-y\|_{2}^{2}
	\label{equ:cd}
\end{equation}

\subsection{Dimensionality Transformation}
To seek optimal and concise features for shape representations, two typical techniques are used to embed shape features in a latent space.
First, Principal components analysis (PCA) \cite{wold1987principal} is selected to provide a sequence of optimal linear transformations for high-dimensional ordered shape features. 
To achieve this goal, PCA computes new variables called \textit{principal components} which are obtained as linear combinations of the original variables. 
Formally, considering a shape feature matrix $\vfX$ with $m$ shapes and $n$ feature dimensions, the goal of PCA is to find a transformation $\vfP$ to linearly convert $\vfX$ to $\vfY$ and reduce the original $n$ feature dimensions into $k$ dimensions ($k<<n$), which can be denoted by $ \vfY = \vfP \vfX $. 
One efficient solution for the PCA problem is known as the singular value decomposition (SVD) \cite{wall2003singular}.
Since semantic analysis of \textbf{LaSeSOM} needs the reconstructed shapes from the low-dimensional latent variables.
For this reason, the inverse sample, $\vfX_{rec}$ reconstructed from the compressed feature is needed, which can be solved by $ \vfX_{rec} =  \vfP^{-1} \vfY + \bm\mu$, where $\bm\mu$ is the mean of normalization.
Besides, to select an appropriate number of components, the explained variance is defined as: $ v_{exp} = {\sum_{i=1}^{k} v_{i}}/{\sum_{i=1}^{n} v_{i}}$.

Second, The auto-encoder (AE) \cite{hinton2006reducing} is used to compress shape features with non-linear transformations. 
Formally, an AE takes an $n$-dimensional soft object shape vector $\vfx$ as its input, which is mapped to its $k$-dimensional bottleneck layer $\vfy$ through the deterministic equation $\vfy=f_{\vftheta}(\vfx)=s(\vfW \vfx+\vfb)$, which in turn is parameterized by $\vftheta=\{\vfW, \vfb\}$. $\vfW$ is a $k \times n$ weight matrix, $\vfb$ is a vector of bias, and $s$ is a $sigmoid$ activation function, $s(x)=\frac{1}{1+e^{-x}}$. The hidden representation is then traced back to a reconstruction $\vfz$ with $n$ dimensions, which is sometimes referred to as the latent representation, where $\vfz=g_{\vftheta^\prime}(\vfy)=s\left(\vfW^{\prime} \vfy+\vfb^{\prime}\right)$, with $\vftheta^{\prime}=\left\{\vfW^{\prime}, \vfb^{\prime}\right\}$. The parameters $\vftheta, \vftheta^{\prime}$ for the model are designed to minimize the average error of reconstruction, which is defined as:
\begin{equation}\small
\begin{aligned} 
\vftheta^{*}, \vftheta^{\prime *}
=& \underset{\vftheta, \vftheta^{\prime}}
{\arg \min } \frac{1}{n} \sum_{i=1}^{n} L\left(\vfx^{(i)}, g_{\vftheta^{\prime}}\left(f_{\vftheta}\left(\vfx^{(i)}\right)\right)\right) 
\label{equ:ae}
\end{aligned}
\end{equation}
where the loss function $L$ needs to be changed depending on the property of input features. 
For example, if the input feature is the ordered features extracted by Fourier descriptor, then $L$ could be normal mean square error (MSE). However, for the unordered point cloud features, the permutation-invariant metric defined in Eq. \ref{equ:cd} is needed to calculate a reconstruction loss.

\subsection{Latent Shape Space}
\begin{figure}[t]
\vspace{0cm}
\centering
\includegraphics[width=1\linewidth]{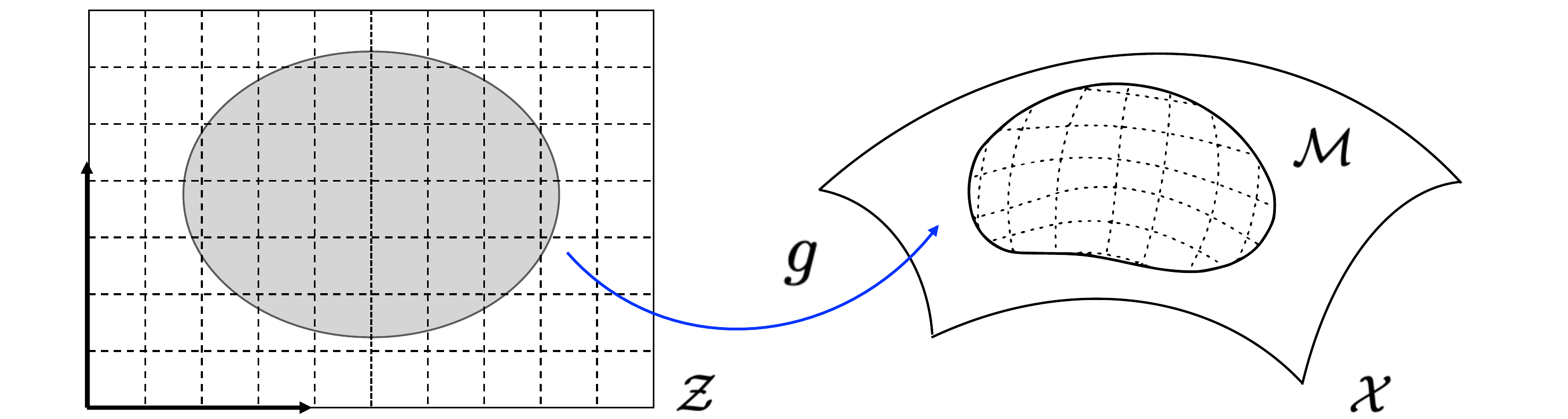}
\DeclareGraphicsExtensions.
\vspace{-0.4cm}
\caption{Conceptual representation of a generator $g$ as a mapping from low-dimensional latent space $\mathcal{Z}$ into a manifold in input data space $\mathcal{X}$.}
\label{fig_LatentMap}
\end{figure}

With dimensionality transformations, we embed the low-level features of the collected shapes in a low-dimensional latent shape space. 
In deep generative models, as shown in Fig. \ref{fig_LatentMap}, a manifold $\mathcal{M}$ is formed through a generator $g$ mapping linear coordinates of variables in latent space $\mathcal{Z}$ ($
\mathcal{Z} \subseteq \mathbb{R}^{k}$) into the curvilinear coordinates of originally high-dimensional shape space $\mathcal{X}$ ($\mathcal{X} \subseteq \mathbb{R}^{n}$, $k \ll n$). Normally, $g$ is a composition function of numerous layers, $g = g^{(1)} \circ g^{(2)} \circ \ldots \circ g^{(\ell)}$, with $\ell$ indexing the layer. 
Combined with a nonlinear activation function $\phi$, it can be represented as below:
\begin{equation}
	g_{k}^{(l)}\left(z^{(l)}\right)=\phi\left(W_{k}^{(l)} z^{(l)}+b^{(l)}\right)
\end{equation}
where $g_{k}^{(l)}$ and $W_k^{(l)}$ denote the $k$th component of the output and $k$th row of the weight matrix, respectively.
The image of $g$ could be a smooth (i.e., $C^\infty$), $k$-dimensional immersed manifold on condition that the Jacobian $J_g(z)$ of $g$ at every point $z \in \mathcal{Z}$ has rank $d$.
According to the chain rules of neural nets, the condition would be satisfied if we choose a smooth and monotonic activation function, $\phi$, and weight matrix has full column rank.
The condition of activation function can be ensured by choosing a correct activation function in the phrase of network construction.
Therefore, $\mathcal{M}$ is a locally differentiable but globally intersected $k$-dimensional Euclidean space (\textit{immersed} manifold).

Mathematically, $\forall z \in \mathcal{Z}$, the Jacobian matrix of $g$,  $J_g(z)$, maps the tangent space of $\mathcal{Z}$ at $z$, $T_z\mathcal{Z}$, to the tangent space of $\mathcal{M}$ at $g(z)$, $T_{g(z)} \mathcal{M}$.
In AE, backpropagation algorithm will calculate out a $k \times n$ partial derivative matrix, $J_g(z)$. 
Consider two vectors $p, q \in T_x \mathcal{M}$ in a linear subspace of $\mathcal{X}$, as a riemannian metric offers the format of an inner product for different tangent vectors in $T_x\mathcal{M}$, therefore, the Riemannian metric of $ \langle u,v \rangle $ can be re-expressed with the dot product of $x$ in the Euclidean space.
Intuitively, the metric denotes the curvature of a Riemannian manifold and measures the extent to which deviates from being Euclidean. 
See standard definitions of Riemannian geometry for a detailed mathematical explaining of curvature \cite{carmo1992riemannian}.



\subsection{Geodesic Path on Manifolds}
Through the mapping $g$, all the concepts (tangent vectors, tangent spaces, curves, etc.) defined in the latent space $\mathcal{Z}$ have an equivalent variable on the manifold $\mathcal{M}$.
For each point $z \in \mathcal{Z}$, the Riemannian metric is defined as below:
\begin{equation}
	G(z)=J_{g}(z)^{T} J_{g}(z)
\end{equation}
Therefore, the inner product of two tangent vectors $u, v \in T_zZ$ is $\langle u, v \rangle = u^T G(z)v$.
Consider a smooth curve in the latent space $\gamma_t: [a,b] \rightarrow \mathcal{Z}$, then it has length $\int_{a}^{b} \left\| \dot{\gamma_t} \right\|  \mathrm{d}t$, where $\dot{\gamma_t} = \mathrm{d} \gamma_{t} / \mathrm{d} t$ denotes the velocity of the curve. The length of this curve $L$ lying on the manifold ($g \circ \gamma(t) \in M$) is computed as:
\begin{equation}
\begin{aligned}
	L\left[g (\gamma_t) \right]
	=\int_{a}^{b} \left\|\dot{g}\left(\gamma_t\right)\right\| \mathrm{d} t =\int_{a}^{b}\left\|\mathbf{J}_{\gamma_{t}} \dot{\gamma}_{t}\right\| \mathrm{d} t	
\end{aligned}
\end{equation}
where $\mathbf{J}_{\gamma_{t}}=\left.\frac{\partial g}{\partial \mathbf{z}}\right|_{\mathbf{z}=\gamma_{t}}$ and the last step follows from Taylor's Theorem, which implies the length of a curve $\gamma_t$ along the surface can be computed directly in the latent space using below defined norm:  
\begin{equation}
\left\|\mathbf{J}_{\gamma} \dot{\gamma}\right\|=\sqrt{\dot{\gamma}^{\top}\left(\mathbf{J}_{\gamma}^{\top} \mathbf{J}_{\gamma}\right) \dot{\gamma}}=\sqrt{\dot{\gamma}^{\top} \mathbf{M}_{\gamma} \dot{\gamma}}
\end{equation}
Here, $\mathbf{M}_{\gamma} = \mathbf{J}_{\gamma}^{\top} \mathbf{J}_{\gamma}$ and it is a symmetric and positive definite matrix, that gives rise to the definition of a Riemannian metric for each point $z$ in the latent space $\mathcal{Z}$. The arc length with metric $\mathbf{M}_\gamma$ can be re-expressed as:
\begin{equation}
	L(\gamma) = \int_{a}^{b} \sqrt{\dot{\gamma}_t^{\top} \mathbf{M}_{\gamma_t} \dot{\gamma}_t} \mathrm{d} t
\end{equation}
To obtain a geodesic curve, the curve length $L(\gamma)$ is locally minimized through an energy functional $E(\gamma)$ defined as:
\begin{equation}
E(\gamma)=\frac{1}{2} \int_{a}^{b} \dot{\gamma}(t)^{T} G_{\gamma(t)} \dot{\gamma}(t) d t
\label{equ:energy}
\end{equation}
In Riemannian geometry, taking a variation of the geodesic energy function can lead to the Euler-Lagrange equation calculated as:
\begin{equation}
\begin{aligned}
	\frac{d^{2} \gamma^{\mu}}{d t^{2}}=-\Gamma_{\alpha \beta}^{\mu} \frac{d \gamma^{\alpha}}{d t} \frac{d \gamma^{\beta}}{d t}	
\end{aligned}
\label{equ:ode}
\end{equation}
where $\Gamma_{\alpha \beta}^{\mu} $ is the Christoffel symbol of the metric $G$, which is defined as:
\begin{equation}
\begin{aligned}
	\Gamma_{\alpha \beta}^{\mu}=\frac{1}{2} G^{v \mu}\left(\frac{\partial G_{v \beta}}{\partial \gamma^{\alpha}}+\frac{\partial G_{v \alpha}}{\partial \gamma^{\beta}}-\frac{\partial G_{\alpha \beta}}{\partial x^{\mu}}\right)
\end{aligned}
\end{equation}
where $G^{v\mu}$ is the inverse of $G_{v\mu}$. 
However, calculation of the Christoffel symbols is considerably expensive, because this process involves the inverse of $G$ and second order derivatives of the $g$. 
Thus, instead of getting the entire geodesic path, we only calculate out few discrete points along on the geodesic path with discrete geodesic energy (\ref{equ:energy}) to avoid expensive calculations.
\begin{algorithm}[htbp]
\label{alg1:geo_path}
\caption{Geodesic Path Generation} 
\LinesNumbered
\KwIn{Two shape coordinates, $z_0, z_N \in \mathcal{Z}$\; learning rate $\alpha \in \mathbb{R}_{+}$ }
\KwOut{discretized geodesic points, $z_0, z_1, \ldots, z_N \in \mathcal{Z}$}
Initialize $z_i$ by a linear interpolation between $z_0$ and $z_N$
\While{$\sum_{i}\left\|\nabla_{z_{i}} E\right\|^{2}>\epsilon$}{
	\For{ $i \in \left\{1, \ldots, N-1 \right\}$}{
		Calculate $\nabla_{z_{i}} E$ using (\ref{equ:graE}) \\
		$z_i \leftarrow z_i - \alpha \nabla_{z_{i}} E$
	}
}
\Return $z_0, z_1, \ldots, z_N$
\end{algorithm}
Formally, consider a discretized curve $\gamma:[0, 1] \rightarrow \mathcal{Z}$ denoted by a series of coordinates $z_{0}, z_{1}, \ldots, z_{N} \in \mathcal{Z}$. With $T$ time steps, a sequence of discrete time intervals, $\delta t = 1 / N$, is generated, which matches a discretized points on the manifold $\mathcal{M}$, $g(z_i)$. With a small shift, the velocity of $g(z_i)$ can be formulated by $v_{i}=\left(g\left(z_{i+1}\right)-g\left(z_{i}\right)\right) / \delta t$. Similarly, the energy of this curve can be given:
\begin{equation}
	E_{z_{i}}=\frac{1}{2} \sum_{i=0}^{N} \frac{1}{\delta t}\left\|g\left(z_{i+1}\right)-g\left(z_{i}\right)\right\|^{2}
\end{equation}
Fixing the first and last points, $z_0$ and $z_N$, as the beginning and ending points of the geodesic curve, minimizing this energy function would result in an approximated geodesic path, which can be obtained by performing a gradient descent algorithm for $z_{1}, \ldots, z_{N-1}$, along this curve. The gradient at $z_i$ is computed as:
\begin{equation}
	\nabla_{z_{i}} E=-\frac{1}{\delta t} J_{g}^{T}\left(z_{i}\right)\left(g\left(z_{i+1}\right)-2 g\left(z_{i}\right)+g\left(z_{i-1}\right)\right)
	\label{equ:graE}
\end{equation} 
Therefore, by implementing a gradient descent algorithm, the calculating process of a discretized geodesic path can avoid the expensive calculations of Christoffel symbols.
The detailed procedures is illustrated in Algorithm \ref{alg1:geo_path}. 
\subsection{Semantic Analysis}
\begin{figure}[t]
\centering{}
\vspace{-0.4cm}
\includegraphics[width=1\linewidth]{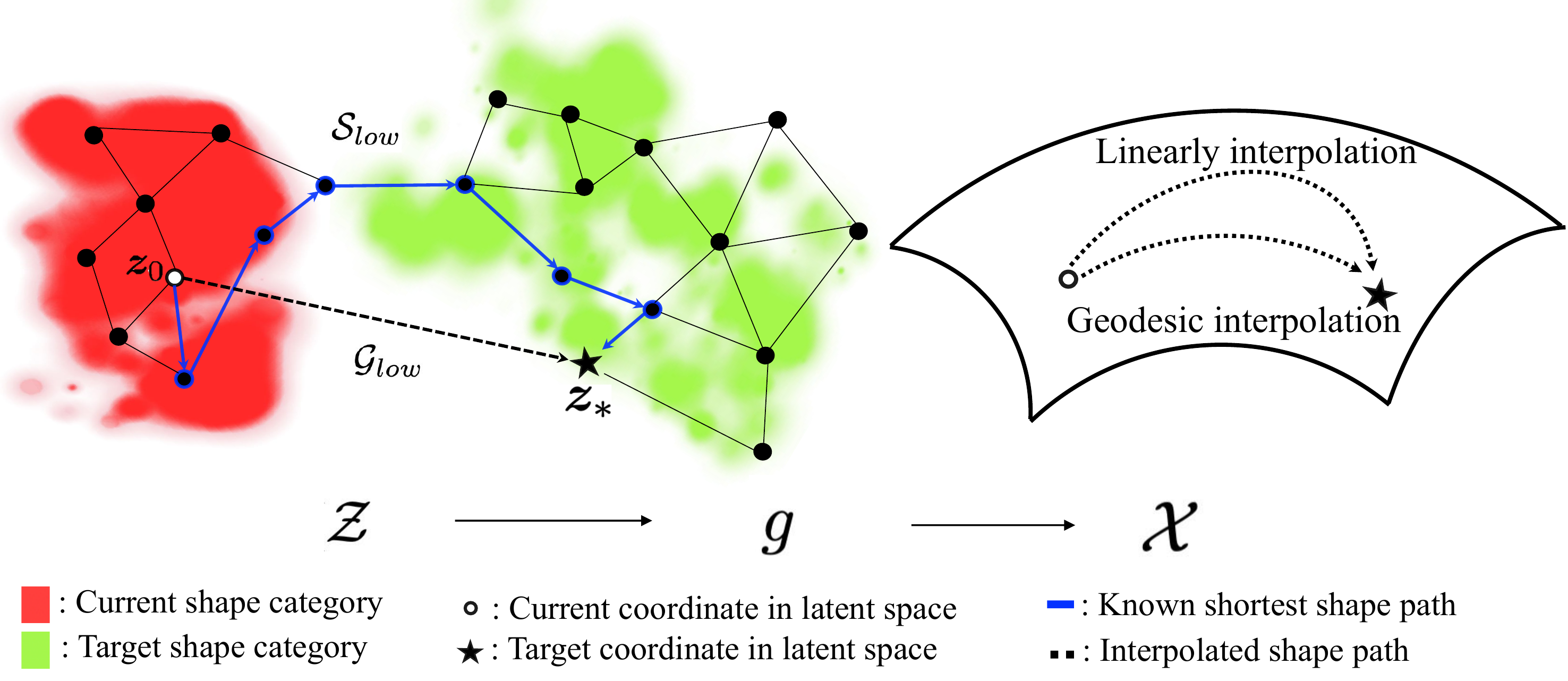}
\DeclareGraphicsExtensions.
\vspace{-0.4cm}
\caption{Depiction of the deformation planning in latent shape space. According to Alg. \ref{alg:shp_pln}, geodesic interpolated path is generated based on the results of linear interpolation in the latent space.
}
\label{fig_shape_query}
\vspace{-0.3cm}
\end{figure}

To make the deformation process of soft objects explainable, semantic analysis techniques are introduced to the high-level representation in \textbf{LaSeSOM}.
First, to identify the effect on each shape dimension, Alg. \ref{alg:fea_sem} is designed. In this algorithm, given a latent variable $\vfz_0$ encoded by function $h$, we gradually increase the $p$-th feature value with a short step $\delta$ for $\vfz_0$ to form a set of changed coordinates, $\mathcal{G}^{(p)}_{low}$, and then we need to update this set based on the whether generator $g$ is not linear. 
At last, we reconstruct the inverse samples $\left\{ \vfx_{1}', \vfx_{2}', \dots, \vfx_{n}' \right\}$ for the soft object. The visualization of these inverse samples allows us to identify the semantic meanings for each dimension of the compressed feature in order to support our high-level semantic shape analysis.
Second, \textit{semantic deformation analysis} is introduced to establish a mapping from soft object deformations to latent variables in latent shape space. Intuitively, 
if the dimensionality reduction technique is invertible, then we can explore deformation rules between different shape classes by observing the latent shape space.  
With performing classification on the latent variables encoded from collected shapes, this path will travel through different spaces enclosed by pre-defined shape classes, thus revealing some rules of shape deformations in real-world applications.
\begin{algorithm}[htbp]
\caption{Semantic Feature Analysis} 
\label{alg:fea_sem}
\LinesNumbered
\KwIn{Shape vector $\vfx_0$, order $p$, step $\delta$, iteration $N$, encoder $h$, decoder $g$}
\KwOut{Semantic deformation trace of $p$-th dim $\mathcal{D}_s^{(p)}$}
Compute the coordinate $\vfz_0$ with $\vfz_0 = h(\vfx_0$) \\
$\mathcal{G}^{(p)}_{low}$ = $\left\{\vfz_0, \vfz_1, \ldots,\vfz_N\right\}$ = Interpolation$(\vfz_0, p, \delta, N)$\\
\If{$g$ \rm is not linear}{
	Update $\mathcal{G}^{(p)}_{low}$ with geodesic Alg. (\ref{alg1:geo_path})
}
$\mathcal{G}^{(p)}_{high} = \left\{ \vfx_{1}', \vfx_{2}', \dots, \vfx_{n}' \right\} = g(\mathcal{G}^{(p)}_{low})$\\
$\mathcal{D}_s^{(p)} = {\rm Visualizer}(\mathcal{G}^{(p)}_{high})$\\
\Return $\mathcal{D}_s^{(p)}$
\end{algorithm}
Third, \textit{latent shape planning} presents a solution to the shape planning problem for soft objects  from the current shape to the target shape. Let the current shape and target shape be  
$\vfx_{0}$ and $\vfx_{*}$, respectively. After dimensionality transformation, the input shapes are transformed to a $k$-dimensional latent shape space  ($
\mathcal{Z} \subseteq \mathbb{R}^{k}$). 
With a encoder $h$, the encoded coordinates of
$\vfz_{0}$ and $\vfz_{*}$ are readily known in this latent space. 
As Fig. \ref{fig_shape_query} shows, shapes are represented as nodes in the latent space and these nodes are connected to form  different neighbor networks with different colors based on the prediction from $k$NN algorithm.
With the implementation of shortest path searching algorithm in the latent shape space, the shape deformation path from the location of current shape to the location of target shape based on the known shape network can be achieved. 
Let $\mathcal{S}_{low}$ denote the shapes lying on the shortest path from $\vfz_{0}$ to $\vfz_{*}$ and let $\mathcal{S}_{high}$ denote the same shape vectors but with high dimensions reconstructed from $\mathcal{S}_{low}$. 
However, $\mathcal{S}_{low}$ can only find out a shortest path built on known shape data set. This latent shape space contains numerous shapes unknown to the dataset. Thus, we first link $\tilde{\vfx}^{\circ}$ to $\tilde{\vfx}^{*}$ with a straight line. Accordingly, $n$ intervals are set to generate $n+1$ intermediate shape statuses denoted by $\mathcal{G}_{low}$ and then could be updated to obtain a shorter geodesic path if the generator is not a linear transformation. 
Note that the linear interpolated path is an intermediate state of geodesic interpolation and they are not exclusive approaches.
 At last, a shape set $\mathcal{G}_{high}$ comprising transitional deformation is formed. Finally, these two deformation paths pass through a visualizer and output the deformation set $\mathcal{D}_{p}$.
\begin{algorithm}[htb]
\caption{Latent Shape Planning} 
\label{alg:shp_pln}
\LinesNumbered
\KwIn{Current shape $\vfx_{0}$, target shape $\vfx_{*}$, iteration $N$, encoder $h$, decoder $g$}
\KwOut{Planned deformation trace $\mathcal{D}_{p}$}
Compute the coordinates using $(\vfz_0, \vfz_{*}) = h(\vfx_0, \vfx_{*})$ \\
$\mathcal{S}_{low}$ = $\left\{\vfz_0, \vfz_1, \ldots,\vfz_*\right\}$ = ShortestPath$(\vfz_0, \vfz_*)$ \\
$\mathcal{S}_{high}$ = $g$($\mathcal{S}_{low}$) \\
$\mathcal{G}_{low}$ = $\left\{\vfz_0, \vfz'_1, \ldots,\vfz_*\right\}$ = Interpolation$(\vfz_0, \vfz_*, N)$\\
\If{$g$  \rm is not linear}{
	Update $\mathcal{G}_{low}$ with geodesic  Alg. (\ref{alg1:geo_path})
}
$\mathcal{G}_{high}$ = $g$($\mathcal{G}_{low}$) \\
$\mathcal{D}_{p}$ = $\left\{ {\rm Visualizer}(\mathcal{S}_{high}), {\rm Visualizer}(\mathcal{G}_{high}) \right\}$\\
\Return $\mathcal{D}_{p}$
\end{algorithm}

\section{Results}\label{sec:results}
In this section, the data collection for building \textbf{LaSeSOM} is described first, and afterwards the framework is used to present different representation results in a robotic teleoperated soft object manipulation task via Leap Motion \cite{jang2019intuitive} demonstration. 

\subsection{Data Collection}
\begin{figure}[t]
\includegraphics[width=1\linewidth]{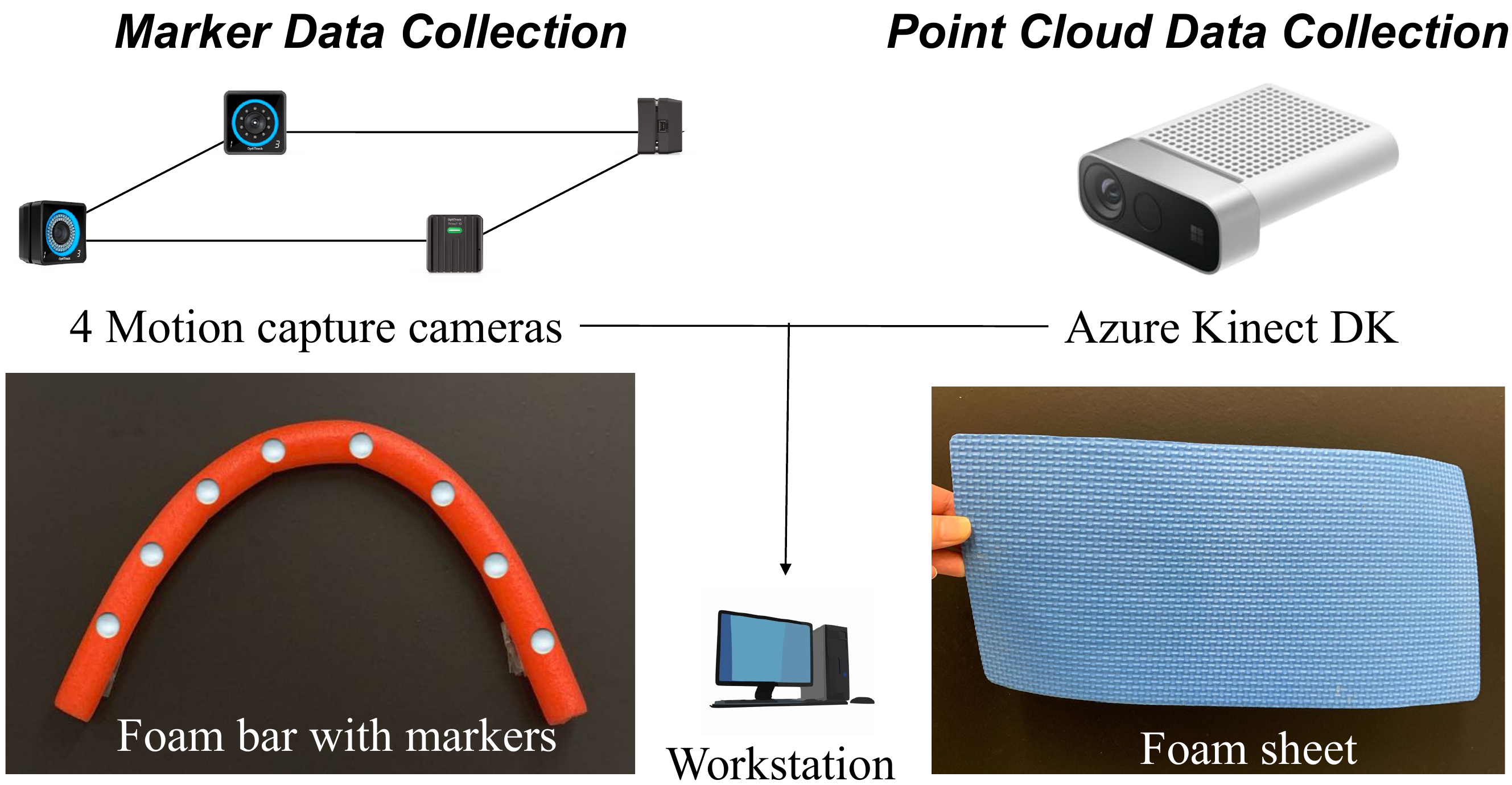}
\DeclareGraphicsExtensions.
\vspace{-0.4cm}
\caption{Experimental setups of the shape data collection to build \textbf{LaSeSOM}. 
Left shows the setup to collect ordered marker data for a foam bar, while right is used to collect unordered point cloud data for a foam sheet.}
\label{fig_collect_setup}
\end{figure}

\begin{figure*}[htbp]
\centering{}
\includegraphics[width=1\linewidth]{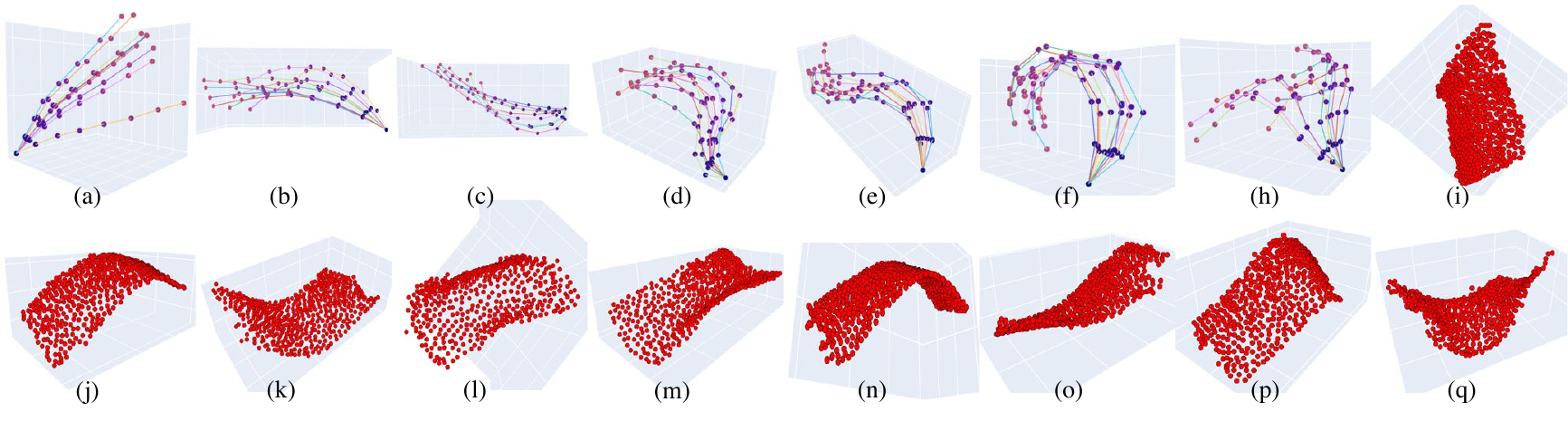}
\DeclareGraphicsExtensions.
\vspace{-0.6cm}
\caption{
Visualizations of shape samples of predefined categories. Figures (a) to (h) shows the seven classes for the foam bar deformation, and figures (i) to (q) presents the nine classes for the foam sheet deformation.
}
\label{fig_cls_disp}
\end{figure*}

As shown in Fig. \ref{fig_collect_setup}, two different soft objects (a foam bar and a foam sheet) were used to collect deformed shapes. 
For the foam bar, the Prime 13 motion tracking system was used to track the position of each marker mounted on the its surface in 30 FPS.
Whereas, the deformations of the foam sheet were captured with a same 30 FPS in a format of point clouds by an RGB-D camera (Azure Kinect DK).
Fig. \ref{fig_cls_disp} displayed few samples for each corresponding categories. Note that the positive and negative categories would be combined or separated based on different analytical needs.

\begin{table}[ht]
\caption{Data Summary}
\label{tab:data_sum}
\vspace{-0.5cm}
\centering
\subfloat
{
\begin{tabular}{llll}
\toprule
Category & Set1 & Set2 \\ [0.5ex] 
\midrule
Line        & 857  & 57  \\ 
Arch Pos.   & 1038 & 825 \\
Arch Neg.   & 1339 & 0   \\
S Pos.      & 1570 & 200 \\
S Neg.      & 1482 & 100 \\
Helix Pos.  & 1005 & 110 \\
Helix Neg.  & 957  & 0   \\
\midrule
\textbf{Total} & 8248  & 1292 \\
\midrule
\end{tabular}

}
\qquad
\subfloat
{
\begin{tabular}{lll}
\toprule
Category & Set \\ [0.5ex] 
\midrule
Plane       & 250 \\ 
Blend \#1 Pos. & 250 \\
Blend \#1 Neg. & 250 \\
Blend \#2 Pos. & 250 \\
Blend \#2 Neg. & 250 \\
Fold \#1 Pos. & 250 \\
Fold \#1 Neg. & 250 \\
Fold \#2 Pos. & 250 \\
Fold \#2 Neg. & 250 \\
\midrule
\textbf{Total} & 2250\\
\midrule
\end{tabular}
}
\vspace{-0.6cm}
\end{table}

\begin{figure*}[htbp]
\centering{}
\includegraphics[width=1\linewidth]{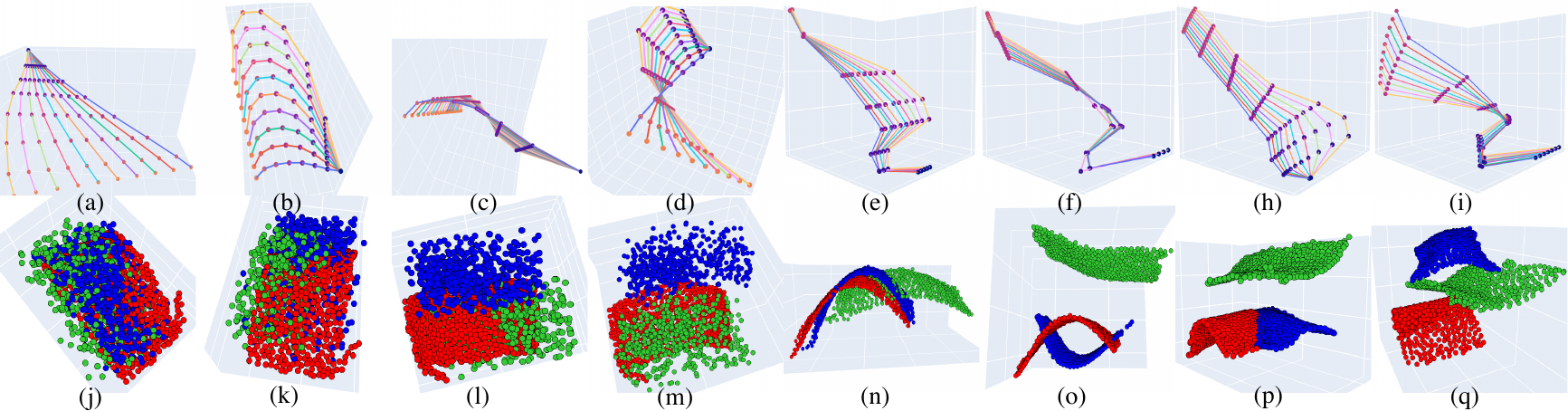}
\DeclareGraphicsExtensions.
\vspace{-0.4cm}
\caption{
Visual comparison of the semantic features from different dimensionality transformation techniques, where figures (a) to (d) and (e) to (i) respectively shows the results of the foam bar from PCA and AE. Figures (j) to (q) shows the visualization results of eight (total in \textit{64-dim}) semantic features.}
\label{fig_sem_fea}
\vspace{-0.5cm}
\end{figure*}
\vspace{-0.2cm}

\subsection{Semantic Feature Analysis}
\subsubsection{Shape Features}
\begin{figure}[htb]
\centering{}
\includegraphics[width=1\linewidth]{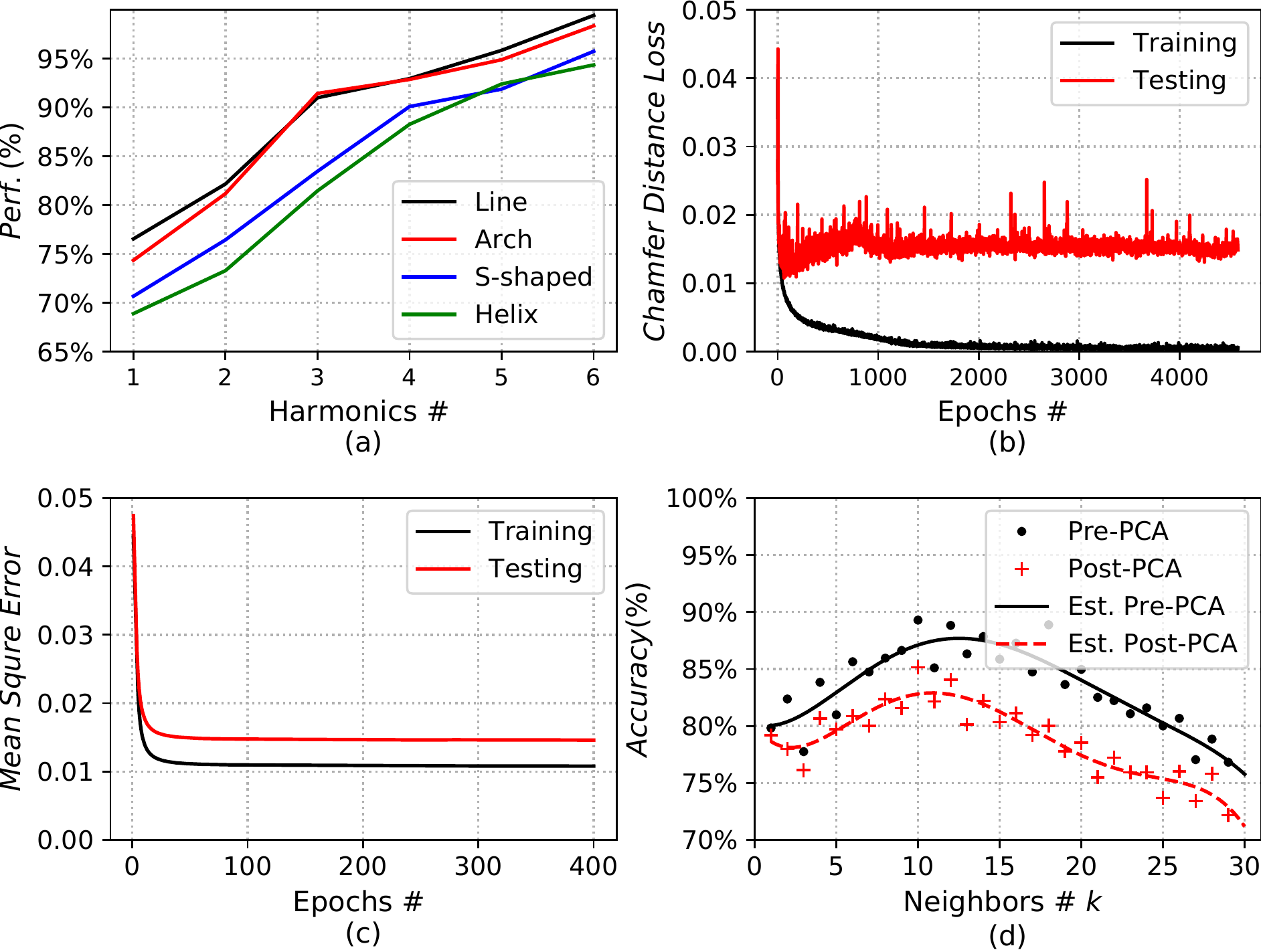}
\DeclareGraphicsExtensions.
\vspace{-0.4cm}
\caption{
(a) The performance of Fourier approximation for four shape classes by different harmonics;
(b) and (c) respectively show the training and validation errors for the corresponding soft objects; (d) presents the pre-PCA and post-PCA classification accuracy for the foam bar. 
}
\label{fig_perf}
\vspace{-0.8cm}
\end{figure}

To examine the fitting performance, the coefficient of determination $R^2$ \cite{nagelkerke1991note}, defined as:
$ 1 - \sum_{i}\left(y_{i}-f_{i}\right)/ \sum_{i}\left(y_{i}-\bar{y}\right)^{2}$, is used to quantify the amount of variability explained by Fourier approximation. 
As shown in Fig. \ref{fig_perf}(a), the shape descriptor becomes more accurate along with the increasing number of harmonics. 
Specifically, the line and arch class shapes demonstrate better performance than the other class shapes under the same number of harmonics, because the S-shaped and helix class shapes are more complex to represent with same number of harmonics.
 
\subsubsection{Reduced Dimensions} 
With PCA performed on Fourier coefficients of marker data, the number of components is set as 4 ( $\text{Var}_{exp} \geq 95\%$ when $k=4$) to investigate following semantic analysis. In the semantic analysis algorithm, parameters are set with iteration $T=10$, $k=4$, and $t=1$ and 
Fig. \ref{fig_sem_fea} (a) to \ref{fig_sem_fea} (d) visually presents the individual semantic effect of the four features. 
Generally, the first component tries to maintain the same shape and alter the angle as the feature value increases, whereas the second component tries to describe the arch shape. The third component is trend to depict the degree of ``S'' shape, whereas the fourth component tries to capture helix shape. 
Though, the results shows partially combined semantic effect (not single effect), each feature dimension has a dominant semantic effect, respectively. Note that the results of the foam sheet with PCA are not presented because PCA can only perform the ordered data.

\begin{table}[ht]
\caption{Network Architecture}
\label{tab:net_struc}
\centering
\vspace{-0.3cm}
\subfloat
{
\scalebox{0.98}{
\begin{tabular}{ll}
\toprule
Marker data (Form Bar) \\ [0.5ex] 
\midrule
\textit{Input 8$\times$3} \\
Flatten \\
FC 8, BatchNorm, ReLU\\ 
FC 4, BatchNorm, ReLU\\ 
FC 8, BatchNorm, Sigmoid\\  
FC 24, Sigmoid\\
Reshape 8$\times$3 \\
\toprule
\end{tabular}}
}
\subfloat
{
\scalebox{0.98}{
\begin{tabular}{ll}
\toprule
Point cloud (Foam Sheet) \\ [0.5ex] 
\midrule
\textit{Input 512$\times$3}\\
3$\times$1 conv, 8, BatchNorm, ReLU\\ 
8$\times$1 conv, 32, BatchNorm, ReLU\\ 
32$\times$1 conv, 64, BatchNorm, ReLU\\ 
Max pool \\
FC 256, Batch norm, Sigmoid\\
FC 512, Batch norm, Sigmoid\\
FC 1536, Sigmoid\\
Reshape 512$\times$3\\
\toprule
\end{tabular}}
}
\vspace{-0.4cm}
\end{table}

To compare with PCA, we implement AE on both marker data and point cloud data with the structure in Tab. \ref{tab:net_struc}.
The latent dimension is kept at 4 for marker's dataset.  
By performing the similar semantic feature analysis on this latent dimensions, Figs. \ref{fig_sem_fea} (e) to \ref{fig_sem_fea} (i) visually present the individual semantic effect of the four dimensions for the code layer. Unlike PCA, these four dimensions mainly depict ``S" shapes from different perspectives, because the neural units in the code layer receive a linear combination from all input data and the S-shaped category accounts for the majority of the training dataset. 
As for point cloud data of the foam sheet, the latent dimension is kept at 64 with the network architecture as shown in Tab. \ref{tab:net_struc}.
and Fig. \ref{fig_perf} (b) shows the corresponding loss trend for training and testing. With the  same implementation of semantic feature analysis, Figs. \ref{fig_perf} (j) to (q) shows the eight reconstructed results out of total the \textit{64-dim} code layer. The red points   represent the raw shape and the blue and green one shows the results of increasing and decreasing feature value, respectively. The former four mainly describe translation of the sheet, whereas the latter four capture the degree of curvature for the foam sheet.
In summary, PCA shows more meaningful semantic analysis results than AE, but it suffers from an unordered data structure. However, AE can perform both ordered and unordered data but hard to explain the semantic meaning of encoded features.
Fig. \ref{fig_perf} (d) shows the best number $k$ for $k$NN to classify the shapes in latent space with a \textit{5-fold} cross-validation and both pre- and post-PCA $k$NN models share a similar trend and reach a peak under the same $k=12$.

\subsection{Latent Shape Space}
\begin{figure}[t]
\centering
\includegraphics[width=1\linewidth]{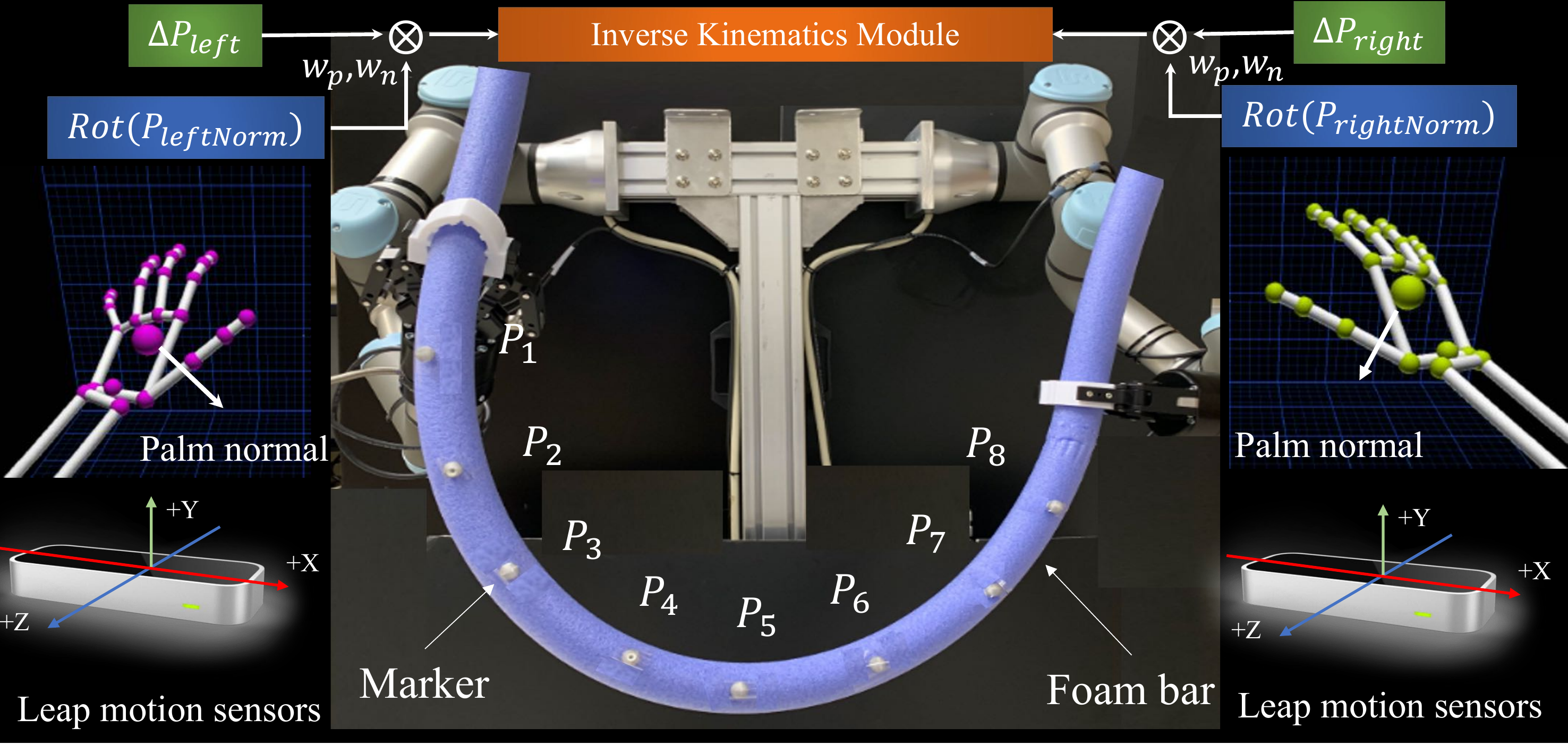}
\DeclareGraphicsExtensions.
\vspace{-0.2cm}
\caption{Architecture of the teleoperated system with Leap Motion sensors for the validation of \textbf{LaSeSOM}.
The new pose is computed with the displacement of palm position $\delta p$ and variance of the hand orientation $Rot(\mathbf{n}_p)$ between $10$ frames, multiplied by appropriate weights $w_p$ and $w_n$.
}
\label{fig_soma_task}
\end{figure}

To imitate the soft object manipulation with human hands, and validate the effectiveness of realtime feedback in a robotic soft object manipulation task, a hand gesture-based teleoperation using Leap Motion \cite{potter2013leap} is an appropriate technique to extract the control signals from hand gestures to teleoperate the soft object in a real-time manner, and the corresponding experimental setup is shown in Fig. \ref{fig_soma_task}, where the robot grippers are fully constrained \cite{dna2014_icra} as during  the data collection stage. The related shape dataset of this manipulation task is shown in the  Table \ref{tab:data_sum} (dataset \#2).

\subsubsection{Semantic Deformation}
As Fig. \ref{fig_shape_plan} (a) shows, all the shapes collected from the foam bar (dataset \#1) are encoded into a 3D latent shape space with $t$-SNE built from AE. 
In this space, the deformation path generated from gesture controls is represented as a red curve and different shape categories of dataset \#1 were organized with \textit{mesh3D} from \textit{Plotly} and rendered with different colors according to the prediction of $k$NN. 
The beginning shape located at the position of the triangle marker, and then the foam bar started from the line category area denoted by a blue color. 
 As the shape deformed, the current  point moved continuously toward the positive arch category denoted by the yellow color in area \#1,
 and then moved to the negative S-shaped category denoted by the cyan color in area \#2. 
 Subsequently, the foam bar went back to the positive arch shape from area \#2 which form a identical but inverse path.
 And so forth, the deformed foam bar ended up with its original shape state. 
Therefore, the entire trace semantically reflects the entire process of shape deformation in a latent space when manipulating a soft object.

\subsubsection{Latent Shape Planning}
\begin{figure}[t]
\centering{}
\includegraphics[width=1\linewidth]{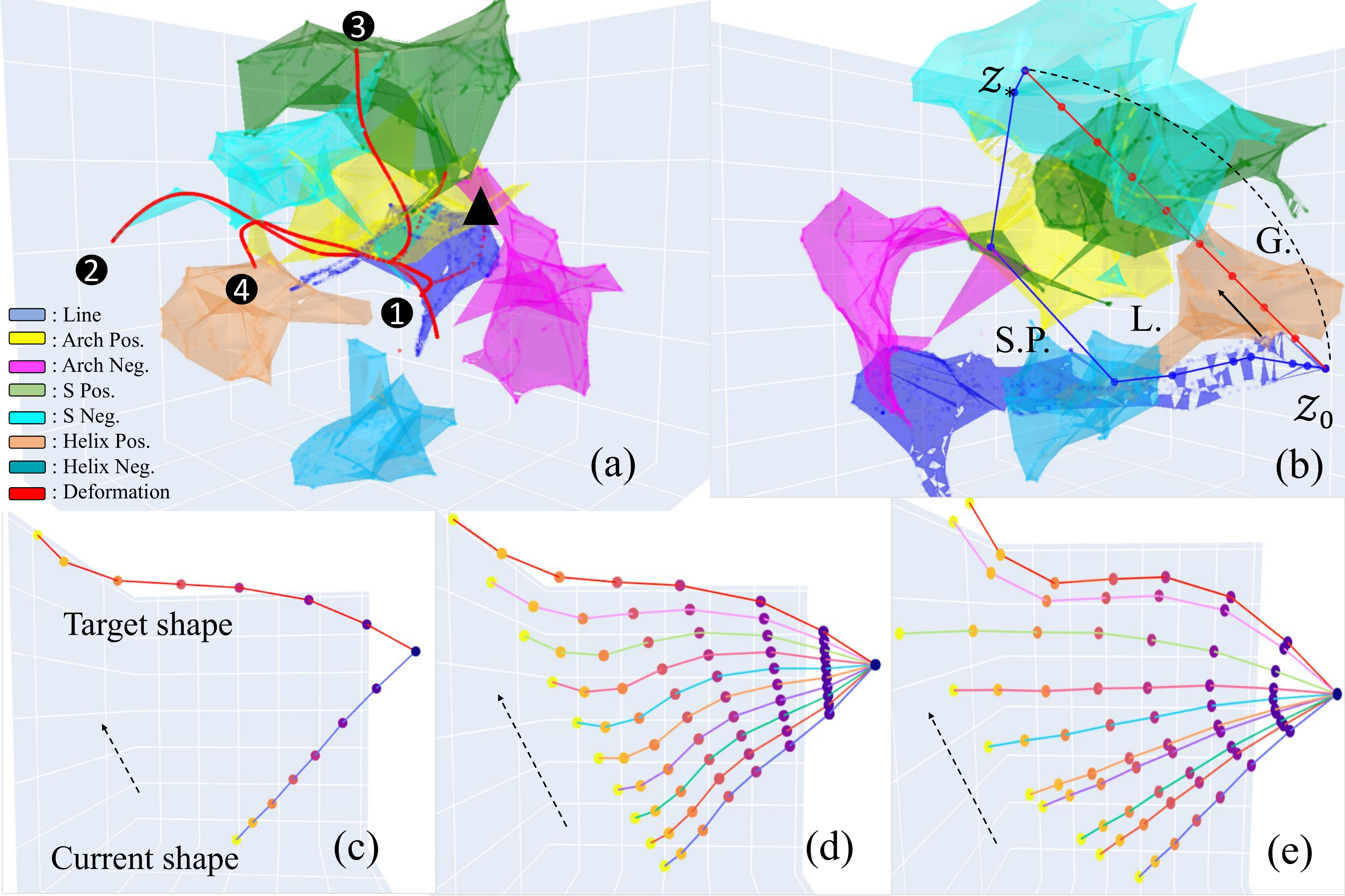}
\DeclareGraphicsExtensions.
\vspace{-0.2cm}
\caption{ Visualization of the process of latent shape planning for the foam bar.
(a) Deformation trace of the manipulation task with Leap motion in latent shape space; (c) shows the beginning shape and the target shape; figures (d) and (e) present the planned shape deformations; (b) presents their corresponding deformation paths with shape planning algorithm. }
\label{fig_shape_plan}
\end{figure}

We use Algorithm 3 to perform a shape planning through a generator ($g: \mathcal{Z} \rightarrow \mathcal{X}$) to map paths calculated in the latent space into shapes on the generated manifold ($\mathcal{M}$). 
Fig. \ref{fig_shape_plan}(b) shows a beginning line and target S-shape of a foam bar. 
With the encoder $h$ (illustrated in Table \ref{tab:net_struc}), we can get encoded shapes in $\mathcal{Z}$ space (see Fig. \ref{fig_shape_plan}-a), which are respectively represented as $\vfz_0$ and $\vfz_*$. 
Then, two sets of shapes are generated based on different calculations in $\mathcal{Z}$ space. 
Shape set $\mathcal{S}_{low}$ denoted by the blue spline is calculated by a shortest path search algorithm on collected data.
In dataset \#1, $\mathcal{S}_{low}$ is a sequence of shape index, 
$ \lbrace x_{540},  x_{532},  x_{530},  x_{526},  x_{568},  x_{777},  x_{774},  x_{1929},  x_{5812},  x_{5040} \rbrace $.
Another shape set $\mathcal{G}_{high}$ is generated by a linear interpolation denoted by the red spline between $\vfz_{0}$ and $\vfz_{*}$ at first, and then an iterative updating on each coordinate with geodesic path illustrated in Alg. \ref{alg1:geo_path}. Figs. \ref{fig_shape_plan} (c) and (d) show the resulting deformation processes from a geodesic interpolation and shortest path, respectively. We can clearly observed that the geodesic path-based interpolation deformation process is smoother compared with the process with a shortest path.

\begin{figure}[ht]
\centering{}
\includegraphics[width=1\linewidth]{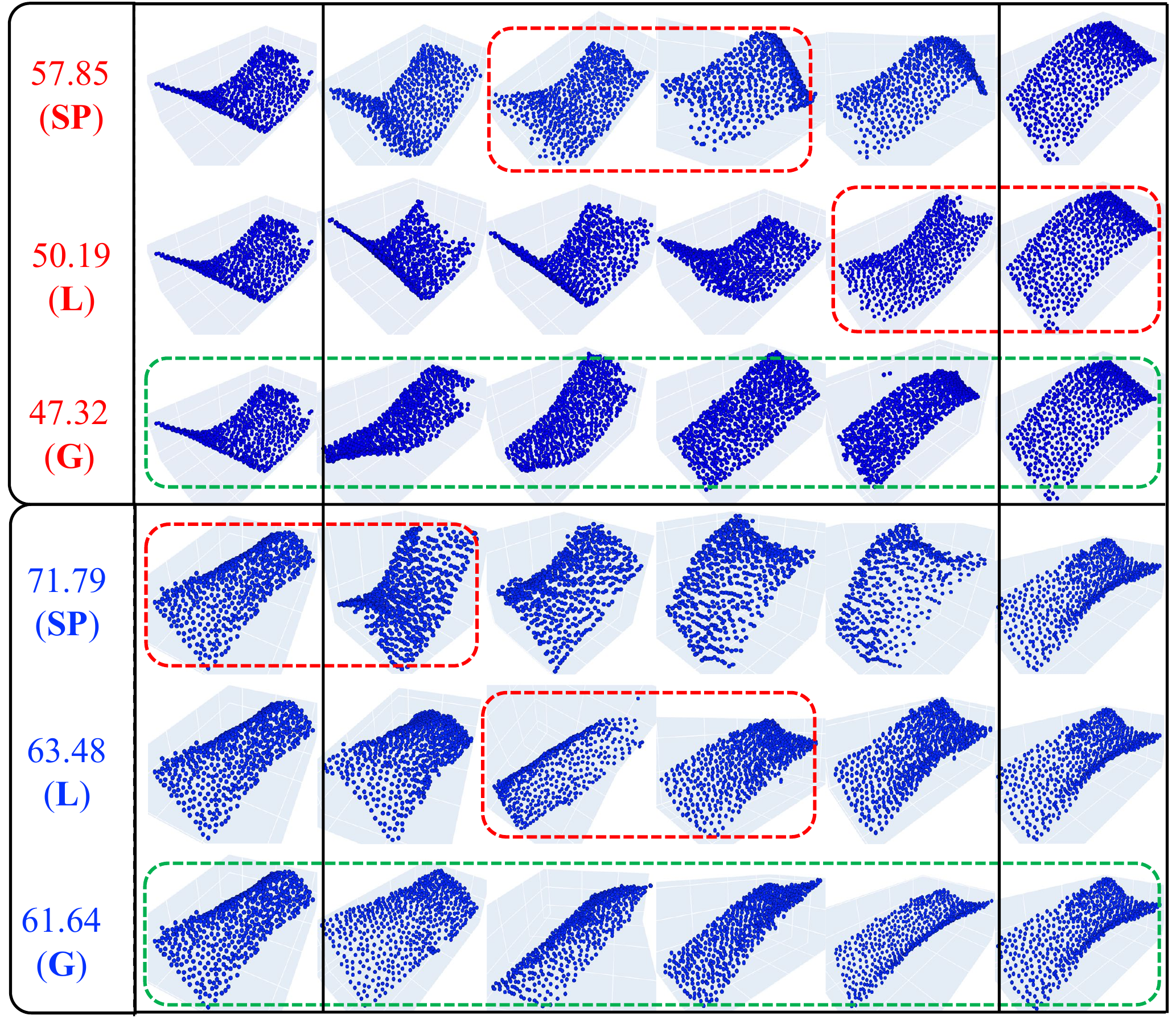}
\vspace{-0.2cm}
\DeclareGraphicsExtensions.
\caption{Shape planning results of shortest path, linear interpolation, and geodesic interpolation for foam sheet dataset. Column 1: arc length; Rows 1, 4: shortest path; Rows 2, 5: linear; Rows 3, 6: geodesic.}
\label{fig13_dist_sheet}
\vspace{-0.2cm}
\end{figure}

To compare geodesic path-based interpolation to its intermediate state (pure linear interpolation), Fig. \ref{fig13_dist_sheet} shows two groups of point clouds (foam sheet) generated with the shortest path, linear interpolation, geodesic interpolation and the corresponding arc lengths. The first column represents the current shapes and the last for the target shapes.
The geodesic path-based interpolation has a shorter arc length on data manifold and smoother morphing process compared with the shortest path and linear interpolation methods, which is supported by the morphing processes marked by green boxes.
In contrast, the shortest path-based and linear interpolation methods show several results (marked by red boxes) with unsatisfied physical feasibility, which may cause excessive stretching and damage the object.
Although, the geodesic curve on the manifold presents a shorter arc length compared with linear interpolation, their difference is not significant, which indicates that the manifold generated by generator architecture for form sheet has little curvature, even non-linear. 

\section{Conclusions}\label{sec:con}
In this paper, we present a generic latent representation framework for semantic soft object manipulation tasks. With dimensionality transformations, we embed the shapes of soft objects from the originally high-dimensional shape space into a semantically low-dimensional latent shape space and solve the shape planning with designed geodesic path-based algorithms on the data manifold. The numerical and experimental results have validated the effectiveness of the proposed framework. As future research, we plan to implement a manipulator with \textbf{LaSeSOM} based feedback control for soft objects and transfer learning for soft object representation. 

\bibliographystyle{IEEEtran}
\bibliography{refs}

\end{document}